\documentclass[letterpaper, 10 pt, conference]{ieeeconf}  

\IEEEoverridecommandlockouts                              

\usepackage{amsmath} 
\usepackage{amssymb}  
\usepackage{algorithm, algpseudocode} 
\usepackage{xcolor}
\usepackage{booktabs}
\usepackage{graphicx}
\usepackage{float}
\usepackage{placeins}
\usepackage{hyperref}
\usepackage[style=ieee]{biblatex}
\usepackage{tikz}
\usepackage{microtype}
\title{\LARGE \bf
Implicit Maximum Likelihood Estimation for Real-time Generative Model Predictive Control
}

\author{%
Grayson Lee, Minh Bui, Shuzi Zhou, Yankai Li, Mo Chen, Ke Li%
\thanks{All authors are with the School of Computing Science, Simon Fraser University, Burnaby, BC, Canada.}%
\thanks{Corresponding author: Grayson Lee, {\tt\small graysonl@sfu.ca}}
}

\begin{document}

\maketitle
\thispagestyle{empty}
\pagestyle{empty}

\setlength{\textfloatsep}{10pt}
\setlength{\floatsep}{10pt}
\setlength{\intextsep}{10pt}
\setlength{\heavyrulewidth}{0.08em}
\setlength{\lightrulewidth}{0.05em}
\setlength{\cmidrulewidth}{0.05em}


\begin{abstract}
Diffusion-based models have recently shown strong performance in trajectory planning, as they are capable of capturing diverse, multimodal distributions of complex behaviors. A key limitation of these models is their slow inference speed, which results from the iterative denoising process. This makes them less suitable for real-time applications such as closed-loop model predictive control (MPC), where plans must be generated quickly and adapted continuously to a changing environment. In this paper, we investigate \emph{Implicit Maximum Likelihood Estimation} (IMLE) as an alternative generative modeling approach for planning. IMLE offers strong mode coverage while enabling inference that is two orders of magnitude faster, making it particularly well suited for real-time MPC tasks. Our results demonstrate that IMLE achieves competitive performance on standard offline reinforcement learning benchmarks compared to the standard diffusion-based planner, while substantially improving planning speed in both open-loop and closed-loop settings. We further validate IMLE in a closed-loop human navigation scenario, operating in real-time, demonstrating how it enables rapid and adaptive plan generation in dynamic environments. Real-world videos and code are available at \url{https://gmpc-imle.github.io/}.
\end{abstract}



\section{INTRODUCTION}

Recent advances in generative modeling, in domains such as image, video, and language, have inspired their application to decision-making problems. In offline reinforcement learning (RL), the objective is to learn an effective policy from a fixed dataset of past interactions without additional environment sampling. One prominent approach is Reinforcement Learning via Supervised Learning (RvS) \cite{emmons2022rvs,schmidhuber2019reinforcement}, which re-frames policy learning as the prediction of actions (or entire action sequences) from states, often conditioned on a desired return. By framing RL problems in the form of supervised learning (SL) the goal is to be able to leverage the success and tools from SL. Generative models offer a natural fit here, as they can capture rich distributions over behaviors and support flexible conditioning on goals, returns, or other task specification.

\begin{figure}[t]
  \hfill
    \includegraphics[width=\columnwidth]{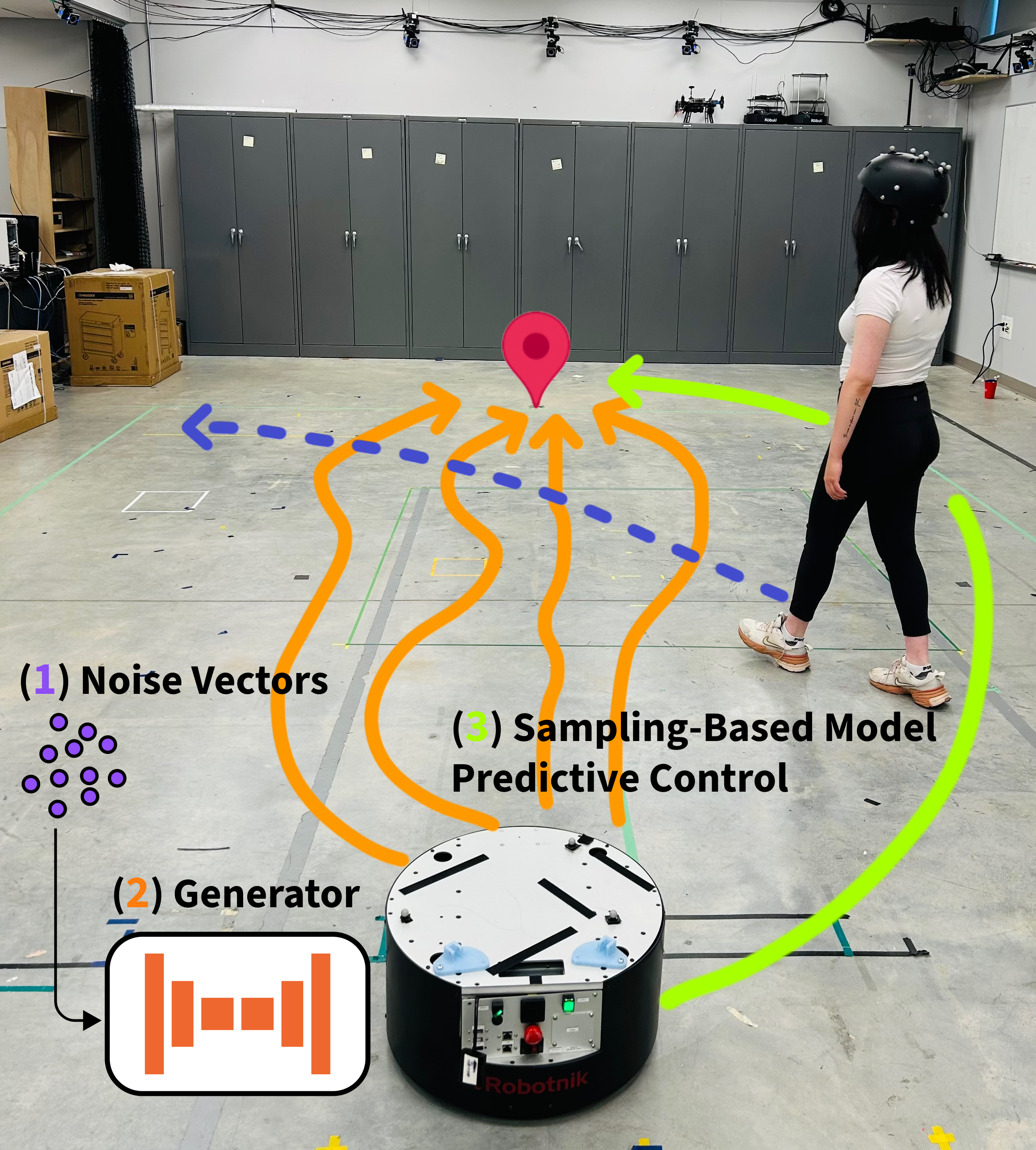}
    \caption{\textbf{IMLE-based planning.} (1) Noise is sampled from a Gaussian distribution. (2) Our model generates a set of candidate trajectories from the sampled noise. (3) Sampling-based model predictive control is used to select and refine the top trajectory.}
    \label{fig:trajectorygeneration}
\end{figure}

One such instance is learning-based trajectory optimization, which directly targets the generation of high-return trajectories. Early approaches include transformer-based models~\cite{chen2021decision,janner2021sequence}, which generate trajectories in an auto-regressive manner; however, these can suffer from compounding errors over long horizons. Diffusion-based planners, such as Diffuser~\cite{janner2022diffuser}, have been proposed to address this limitation. These methods formulate trajectory generation as a denoising process: starting from noise, they iteratively refine samples to match the distribution of high-return trajectories. Operating directly at the trajectory level improves temporal coherence and mitigates compounding errors, enabled by their ability to capture rich, multimodal trajectory distributions. However, a key drawback is the slow inference speed due to the iterative nature of the sampling process. Recent work explores hierarchical modeling~\cite{dong2024diffuserlite} and policy distillation~\cite{lu2025habitizing} to mitigate this bottleneck and make diffusion-based planning feasible for real-time or robotics applications, but introduce additional complexity and can be hard to train in practice.

\textit{Implicit Maximum Likelihood Estimation (IMLE)}~\cite{li2018implicit,Aghabozorgi2023AdaptiveIF,vashist2024rejection} offers a promising alternative. Unlike diffusion models, IMLE generates samples in the same way as GANs~\cite{goodfellow2014generative} with a single forward pass, avoiding the computational burden of iterative denoising. In contrast to the popular GANs, which suffer mode collapse, IMLE exhibits strong mode coverage guarantees due to its loss objective. These properties make IMLE a compelling candidate for trajectory-level generative modeling and planning. In this work, we present an alternative to diffusion-based planners, which is conceptually simple and enables fast inference speed, while rivaling the performance of diffusion-based methods. Moreover, we show it is extendable to real-time applications.
\begin{itemize}
    \item We propose a trajectory generation framework based on Implicit Maximum Likelihood Estimation (IMLE), adapted for conditional generative modeling in planning domains.
    \item Our approach supports single-shot trajectory sampling, avoiding the computational cost of iterative inference required by diffusion-based methods.
    \item We show that our method achieves competitive planning performance while significantly improving inference speed on various Mujuco benchmarks.
    \item We demonstrate real-time path planning of our method on a mobile robot navigating among humans in the real world, demonstrating its capacity for fast and reactive planning.
\end{itemize}

\section{RELATED WORK}

\subsection{Diffusion for Planning}
Diffusion models for planning~\cite{janner2022diffuser,ajay2023is,xiao2023safediffuser} treat a trajectory analogously to an image, where the temporal horizon defines the width and the state–action dimensions define the height. The forward process gradually corrupts a clean trajectory $\tau^0$ by adding Gaussian noise at each step:
\begin{equation}
q(\tau^t \mid \tau^{t-1}) = \mathcal{N}\!\left(\tau^t; \sqrt{\alpha_t}\,\tau^{t-1}, (1-\alpha_t) I \right),
\end{equation}
where $\beta_t \in (0,1)$ denotes the noise variance at step $t$, and we define $\alpha_t = 1 - \beta_t$ with cumulative product $\bar\alpha_t = \prod_{i=1}^t \alpha_i$.  
The closed-form forward distribution after $t$ steps is:
\begin{equation}
q(\tau^t \mid \tau^0) = \mathcal{N}\!\left(\tau^t; \sqrt{\bar\alpha_t}\,\tau^0, (1-\bar\alpha_t) I \right).
\end{equation}
As $T$ increases, $\bar\alpha_T$ decays toward zero, and $\tau^T$ converges in distribution to an isotropic Gaussian $\mathcal{N}(0, I)$.

The reverse process is then modeled as a Gaussian denoiser,
\begin{equation}
p_\theta(\tau^{t-1} \mid \tau^t) 
= \mathcal{N}\!\left(\tau^{t-1}; \mu_\theta(\tau^t,t), \Sigma^t \right),
\end{equation}
with the mean $\mu_\theta$ learned by a neural network and a covariance matrix $\Sigma^t$. Prior work in planning \cite{janner2022diffuser,ajay2023is,xiao2023safediffuser} follow the Denoising Diffusion Probabilistic Model (DDPM) framework \cite{ho2020denoising}, which trains a neural network $\epsilon_\theta$ to predict the injected noise by minimizing:
\begin{equation}
\label{eq:DDPM_loss}
\mathcal{L}_{\text{DDPM}} = \mathbb{E}_{\tau^0, \epsilon, t} \Big[ \big\| \epsilon - \epsilon_\theta\!\left( \sqrt{\bar\alpha_t}\,\tau^0 + \sqrt{1 - \bar\alpha_t}\,\epsilon, t \right) \big\|^2 \Big],
\end{equation}
where $\tau^0$ is a trajectory from the dataset and $\epsilon \sim \mathcal{N}(0, I)$.
At inference, trajectories are generated by denoising from $\tau^T$ to $\tau^0$, conditioned on an initial state $s_0$ and/or goal states. To bias generation toward high-return behaviors, classifier-based guidance~\cite{dhariwal2021diffusion} is applied during sampling. Thus, a limitation of this approach is the inference speed, as the denoising process requires many iterative steps, when compared with single-shot generative methods.



\subsection{Diffusion Models for Safe Navigation}
A growing body of work has explored the combination of generative models with control-theoretic tools to achieve safe decision-making in dynamic environments. One of the first examples is Safe Diffuser \cite{xiao2023safediffuser}, which augments the diffusion denoising process with a CBF constraint \cite{nguyen2016exponential, CBForiginal} which they enforce while staying close to the original diffusion step. This work established the basic principle of coupling denoising with explicit safety mechanisms.

The work most directly related to ours in safe navigation is CoBL-Diffusion \cite{mizuta2024cobl}, which integrates Control Barrier Functions (CBFs) and Control Lyapunov Functions (CLFs) into a diffusion-based planning framework. CoBL-Diffusion generates only actions which they then feed through a predefined dynamics to get, and uses CBF/CLF rewards as their guidance function for the denoising process. This ensures consistency between control inputs and resulting states, allowing the diffusion sampler to respect both safety and goal-reaching requirements in dynamic multi-agent environments. However, diffusion-based methods remain computationally expensive and struggle to achieve real-time performance due to their iterative sampling procedure.  

A subsequent line of work replaces diffusion with Conditional Flow Matching (CFM) \cite{mizuta2025unified}, where reward gradients are injected directly into the ODE dynamics to bias the generative process toward safe trajectories. This significantly reduces inference time compared to diffusion, though the procedure remains iterative, however with fewer steps. In contrast, we propose a single-shot IMLE planner that enforces mode coverage and integrates constraints directly into the loss, avoiding the iterative process altogether.

\section{BACKGROUND}
We follow the formulation of \cite{janner2022diffuser}, which models entire trajectories directly with a generative model, as opposed to generating them step-by-step in an autoregressive manner, enabling global consistency and flexible conditioning at the trajectory level. 

\subsection{Problem Setting}
We consider trajectory optimization in the offline setting.
Let $\tau=(s_0,a_0,r_0,\ldots,s_T,a_T,r_T)$ denote a trajectory. We denote the dataset by $\mathcal D=\{\tau_i\}_{i=1}^N$,
collected from prior interactions with the environment.

In optimal control theory, the main objective is to find a sequence of actions that maximizes the cumulative return for a horizon $T$:
\begin{equation}
\label{eq: cumulative_rewards}
a^*_{0:T} = \arg\max_{a_{0:T}} \sum_{t=0}^{T} r(s_t,a_t).
\end{equation}
Generally this optimization is challenging to be solved in high dimensions due to the curse of dimensionality and the complexity of nonlinear dynamics \cite{kappen2005path}.

A common probabilistic relaxation is provided by stochastic optimal control (path-integral / KL-control) \cite{kappen2005path, todorov2009compositionality,williams2016aggressive} and control-as-inference (CAI) \cite{ziebart2010modeling, levine2018reinforcement}. For temperature $1/\beta > 0$ this yields the Gibbs distribution
\begin{equation}
p^{*}(\tau \mid s_0) \;\propto\; p_0(\tau \mid s_0)\,
\exp\Big(\beta \cdot \sum_{t=0}^T r(s_t,a_t)\Big),
\end{equation}
where $p_0(\tau \mid s_0)$ denotes the passive/base dynamics. 
This distribution is the \emph{optimal solution to the entropy-regularized control problem}; as $\beta \to \infty$ (zero-temperature limit), it concentrates on the original argmax solution. 

In practice for our problem setting, we only have access to an offline dataset $\mathcal{D}$, which provides samples from a behavior distribution that may differ significantly from $p^{*}(\tau \mid s_0)$. 
Prior work \cite{janner2022diffuser,ajay2023is} trains a generative model of trajectories (approximating the behavior distribution) and applies energy-based guidance at inference using a learned reward function $r_\theta(\tau) \approx \sum_{t=0}^T r(s_t,a_t)$. Such guidance has been shown to bias sampling toward higher-return trajectories and thereby provide a rough approximation to the CAI formulation, with the behavior distribution acting as the base \cite{lu2023contrastive,feng2025on}.

\subsection{Implicit Maximum Likelihood Estimation}
Implicit Maximum Likelihood Estimation (IMLE)~\cite{li2018implicit,Aghabozorgi2023AdaptiveIF,vashist2024rejection} is a method for training implicit generative models by directly encouraging coverage of the data distribution. The approach uses a generator network $f_\theta$ that maps latent codes $z$, sampled from a standard normal distribution, to generated samples $y = f_\theta(z)$. Rather than requiring an explicit likelihood or adversarial objective, IMLE optimizes the generator to ensure that, for every data point in the training set, there exists a latent code such that the generated output is close to that data point. The method requires sampling at least as many latent codes as there are data points, i.e., $m\geq N$.
To differentiate between the latent pools for unconditional and conditional IMLE, we write $\mathcal{Z}:=\{z^{(j)}\}_{j=1}^m$ for a global latent pool and $\mathcal{Z}_i:=\{z_i^{(j)}\}_{j=1}^m$ for a per-context pool, where $z^{(j)}, z_i^{(j)} \sim \mathcal{N}(0,I)$.

\begin{equation}
\label{eq:imle}
\min_\theta\;
\mathbb{E}_{\mathcal{Z}}
\Big[\,\sum_{i=1}^N \min_{z\in\mathcal{Z}}\,
d \big(f_\theta(z),\,\tau_i\big)\,\Big].
\end{equation}
where $d(\cdot,\cdot)$ is a distance metric over trajectories and $m$ is the total number of latent codes drawn.

In the conditional setting, Conditional IMLE (cIMLE)~\cite{li2020multimodal,peng2022chimle,rana2025imle} extends this idea by training a conditional generator $f_\theta(z, c)$, adding $c$ as a conditioning variable (e.g., the initial state or task specification). The generator maps $(z, c)$ to a sample $\hat{\tau} = f_\theta(z, c)$, and is trained to ensure that for every ground-truth data point $\tau_i$ in the dataset, there exists some latent code $z_i^*$ such that $f_\theta(z_i^*, c_i) \approx \tau_i$.

The cIMLE training objective is defined as:
\begin{equation}
\label{eq:cimle}
\min_\theta\;
\mathbb{E}_{\{\mathcal{Z}_i\}_{i=1}^N}
\Big[\,\sum_{i=1}^N \min_{z\in \mathcal{Z}_i}\,
d\big(f_\theta(z,\,c_i),\,\tau_i\big)\,\Big].
\end{equation}
where $m$ is the total number of latent codes drawn. In the unconditional setting, these $m$ samples are generated once and each data point finds its closest match among them. In conditional IMLE, we generate $m$ samples for each conditioning context $c_i$.

\begin{figure}[t]
  \centering
  \includegraphics[width=\linewidth]{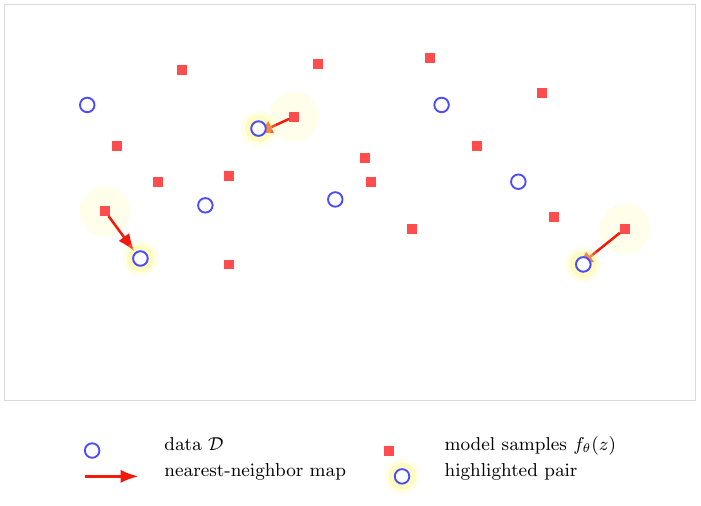}
  \caption{Illustration of IMLE's nearest neighbor matching. The method enforces that, for every data point, there exists a generated sample in its neighborhood, ensuring that the generator covers the full data distribution.}
  \label{fig:imle-nn}
\end{figure}

During training, for each data point $(\tau_i, c_i)$, the model samples a fixed number of latent codes $\{z_i^{(j)}\}_{j=1}^m$, generates corresponding predictions $\{f_\theta(z_i^{(j)}, c_i)\}$, and selects the sample that is closest (under $d$) to the ground-truth trajectory $\tau_i$ as illustrated in Fig.~\ref{fig:imle-nn}. The generator parameters $\theta$ are then updated to minimize this minimum-distance loss.

\section{PLANNING WITH IMLE}

We propose a generative planning framework based on Implicit Maximum Likelihood Estimation (IMLE), designed to avoid the iterative sampling process of diffusion-based methods and produce trajectory candidates in a single forward pass. By leveraging the mode-covering behavior of IMLE, our method can produce high-quality and diverse trajectory candidates for model predictive control (MPC). Moreover, we propose a way of leveraging reward weighting in the context of IMLE model training to enable effective planning even under mixed-quality data.

\subsection{IMLE for Trajectory Generation}

Given a dataset of trajectories $\{\tau_i\}_{i=1}^{N}$, our goal is to learn a conditional generative model $f_\theta(z, c)$ that maps latent codes $z \sim \mathcal{N}(0, I)$ and context $c$ (e.g., initial state, goal state) to full trajectories. IMLE ensures that for each data point $\tau_i$, there exists a latent code $z_i^*$ such that the generated sample $f_\theta(z_i^*, c_i)$ closely matches the ground-truth trajectory $\tau_i$. For our distance metric we choose $\ell^2$-norm which is a standard metric for trajectory generation \cite{janner2022diffuser}. Formally, we minimize the following objective:
\begin{equation}
\mathcal{L}_{\text{IMLE}}(\theta) = \mathbb{E}_{\{\mathcal{Z}_i\}_{i=1}^N}\left[\sum_{i=1}^{N} \min_{z \in \mathcal{Z}_i} \|f_\theta(z, c_i) - \tau_i\|_2^2\right].
\end{equation}
This training paradigm ensures that the learned model covers all modes of the trajectory distribution.

Unlike diffusion models, which rely on a multi-step denoising process, IMLE learns through a direct reconstruction loss. This objective is simpler, requiring no noise scheduling, score matching, or iterative refinement, while still enforcing coverage of the full data distribution \cite{li2018implicit}. As a result, IMLE enables efficient rollout sampling for closed-loop planning. Because IMLE does not support iterative classifier or reward guidance, we instead bias generation toward high-reward trajectories through a modified training loss.

\vspace{1mm}
\begin{algorithm}[ht]
\caption{Reward-Weighted cIMLE Training}
\label{alg:rw-imle}
\textbf{Input:} Dataset $D=\{(\tau_i, c_i, r_i)\}_{i=1}^N$, generator $f_\theta(z,c)$, sample factor $m$, epochs $K$, inner steps $L$, step size $\eta$ \\
\textbf{Precompute:} $w_i \!\gets\! \exp\!\Big(\tfrac{r_i - \mathrm{median}(r)}{\beta \cdot \mathrm{MAD}(r)}\Big)$ or
$w_i \!\gets\! \frac{r_i - r_{\min}}{\,r_{\max}-r_{\min}}$
\begin{algorithmic}[1]
\For{$k=1$ to $K$}
  \State Sample batch $S \subseteq [N]$
  \State Sample latent pools $ \{\mathcal{Z}_i\}_{i=1}^N \sim \mathcal{N}\!(0,I),\;\;\forall i \in S$
  \State $z_i^\star \gets \arg\min_{z\in\mathcal{Z}_i} \|f_\theta(z,c_i)-\tau_i\|_2^2,\;\; \forall i \in S$
  \For{$\ell=1$ to $L$}
    \State Sample mini-batch $\tilde S \subseteq S$
    \State $\displaystyle 
       \theta \gets \theta - \eta \,\nabla_\theta 
       \frac{N}{|\tilde S|} \sum_{i\in\tilde S} 
         w_i \,\|f_\theta(z_i^\star,c_i)-\tau_i\|_2^2$
  \EndFor
\EndFor
\State \Return $\theta$
\end{algorithmic}
\end{algorithm}

\subsection{Reward-Weighted IMLE}
To achieve this bias toward higher-quality trajectories, we modify the IMLE objective to approximate the CAI target distribution $p^*(\tau \mid s_0) \propto p_0(\tau \mid s_0)\exp(R(\tau))$ from Section III. Since this distribution naturally assigns higher probability to high-return trajectories, we incorporate this weighting directly into our learning objective by reweighting each training example according to its return.

Let $r_i=R(\tau_i)$ denote the return of the trajectory $\tau_i$, where $R(\cdot)$ is the reward or cost function for a given task (e.g., environment reward in offline RL benchmarks, or CBF-based safety cost in navigation). We introduce per-sample weights $w_i$ as a function of $r_i$ and define the reward-weighted variant as:
\begin{equation}
\mathcal{L}_{\text{weighted}}(\theta)
\;=\;
\mathbb{E}_{\{\mathcal{Z}_i\}_{i=1}^N}\left[\sum_{i=1}^{N} w_i \cdot
\min_{z\in \mathcal{Z}_i} \,\big\| f_\theta(z, c_i)-\tau_i \big\|_2^2
\right].
\end{equation}

\paragraph{Exponential (Boltzmann) weights}
Following CAI, we use an exponential weighting with robust centering and scaling inspired by \cite{peng2019advantage}:
\[
w_i
\;=\;
\exp\left(\frac{r_i - \mathrm{median}(r)}{\beta\cdot \mathrm{MAD}(r)}\right),
\]
where $\beta>0$ is a temperature parameter.

\paragraph{Linear weights}
We construct a simple linear weighting scheme to compare against:
\[
w_i \;=\; \frac{r_i - r_{\min}}{\,r_{\max}-r_{\min}}.
\]
This baseline is included to verify that exponential weighting gains stem from its CAI grounding rather than a generic preference for high-return samples.





\subsection{Architecture and Conditioning}

Our generative architecture follows a U-Net backbone, as commonly used in trajectory generation \cite{janner2022diffuser}. Diffusion-based methods typically condition trajectories via inpainting, fixing the start and goal states throughout the denoising process to enforce consistency. As our model generates trajectories in a single shot, we instead use Feature-wise Linear Modulation (FiLM) \cite{perez2018film} for conditioning, as recently applied in diffusion-based policy learning \cite{chi2023diffusion}. FiLM injects the conditioning signal at each layer of the network:
\[
\text{FiLM}(x; c) = \gamma(c) \cdot x + \beta(c),
\]
where $\gamma(c)$ and $\beta(c)$ are produced by MLPs applied to the conditioning input $c$.

\begin{figure}[H]
  \centering
  \includegraphics[width=\linewidth]{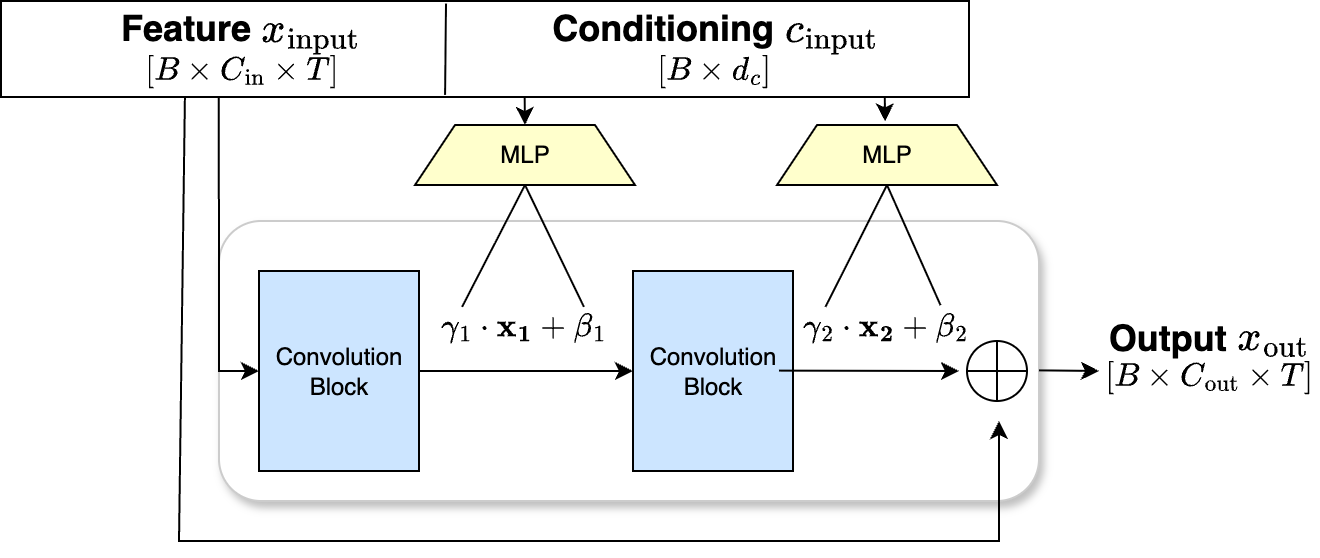}
    \caption{FiLM-conditioned block used throughout the U-Net.
    Two small MLPs conditioning $c$ to scale $\gamma(c)$ and shift $\beta(c)$, which modulate the two blocks via FiLM ($x \mapsto \gamma(c)\odot x + \beta(c)$).}
  \label{fig:FiLM}
\end{figure}
This architecture ensures that the conditioning signals (initial and goal states) are available at all spatial-temporal scales of the network.

\subsection{Applications in Model Predictive Control}
We integrate our IMLE planner into different sampling-based model predictive control (MPC) frameworks to highlight its versatility across domains. In all cases, IMLE provides diverse candidate trajectories in a single forward pass,  which we treat as a learned base distribution.

\paragraph{Score-Ranked MPC} 
On offline reinforcement learning benchmarks~\cite{fu2020d4rl} (e.g., D4RL locomotion), we follow the score-and-ranking sampling-based MPC procedure used in prior work \cite{janner2022diffuser}. At each step, the planner generates a batch of candidate trajectories. These are evaluated with a learned reward function, and the top-ranked trajectory is executed. This setup ensures fairness against diffusion-based planners while emphasizing the computational efficiency of IMLE. 

\paragraph{Model Predictive Path Integral (MPPI)} 
Due to the dynamic nature of pedestrian navigation, we integrate IMLE within the Model Predictive Path Integral (MPPI) framework~\cite{williams2016aggressive}. IMLE replaces the standard Gaussian proposal distribution with a learned multi-modal trajectory generator, producing structured rollouts under tight latency constraints. 

The control objective combines a CBF safety penalty (with adjustable radius at inference) and a CLF goal-progress term~\cite{mizuta2024cobl}, along with a temporally discounted penalty on deviations from the previous plan to reduce oscillations from mode switching, since IMLE generates diverse trajectories.

\section{EXPERIMENTS}

We assess the proposed IMLE-based framework across a diverse set of domains, spanning both offline reinforcement learning and real-time control tasks. Our goals are twofold: (i) to measure performance on established offline RL benchmarks, and (ii) to validate responsiveness in a dynamic, real-time environment.

Our experiments are organized into two categories. In \emph{Offline Reinforcement Learning}, we benchmark on the D4RL MuJoCo locomotion suite~\cite{fu2020d4rl} to study return maximization under various data distributions, and on Maze2D to test long-horizon planning under sparse rewards. Following the setup described in Section IV-D-a, we use a score-ranked MPC for locomotion tasks. For Maze2D, where demonstrations are already successful, we omit the ranking and simply execute a sampled trajectory in an open-loop evaluation. We compare against Diffuser \cite{janner2022diffuser}, which provides a natural baseline, given we use the same UNet architecture and its role as the standard diffusion-based trajectory planner. We measure the sampling frequency on a CPU (AMD EPYC 9655 - Zen 5) and on a GPU (2/7th of NVIDIA H100 as a MIG instance), with a 20 GB memory allocation. All experiments are repeated over 150 random seeds to ensure statistical robustness and align with prior works.

In \emph{Real-Time Navigation}, we evaluate both open-loop simulation and closed-loop deployment on a mobile robot in pedestrian environments. Our planner embeds IMLE within MPPI using the control cost described in Section~IV-D-b.

All models are trained on ETH~\cite{Pellegrini2009youll} and UCY~\cite{Lerner2007crowds} pedestrian trajectories processed with TrajData~\cite{ivanovic2023trajdata}, using horizon $20$ and discretization $0.4\,\mathrm{s}$. Safe behavior is encouraged via reward weighting with a Control Barrier Function (CBF) under a conservative $1\,\mathrm{m}$ safety constraint, and we evaluate collision radii of $0.5$ and $0.7\,\mathrm{m}$. Because few trajectories satisfy this constraint, we augment the dataset through translation, rotation, and smoothing. Pedestrians are modeled as constant-velocity obstacles, and collisions are evaluated against ground-truth trajectories.

For simulation, we train on the augmented ETH dataset and evaluate on 500 UCY scenes. IMLE-MPPI is compared against Gaussian MPPI~\cite{williams2016aggressive} (warm-started with a straight-line trajectory), diffusion-based planners~\cite{mizuta2024cobl}, and flow matching methods. Metrics include collision rate (any collision within radius $r$), goal error (final distance to the goal), smoothness (maximum per-step change in velocity), and jerk (mean magnitude of the second finite difference of velocity).

For real-world experiments, we train on the combined augmented ETH and UCY datasets and deploy the policy on a mobile robot. Diffusion-based planners are excluded due to the latency of iterative denoising, which prevents real-time inference, so we focus on IMLE-based trajectory proposals within MPPI.

Although pedestrian datasets do not perfectly match robot dynamics, incorporating robot dynamics into reward weighting or MPPI refinement is left for future work.

\subsection{Offline Reinforcement Learning}
\begin{figure}[t]
    \includegraphics[width=.95\columnwidth]{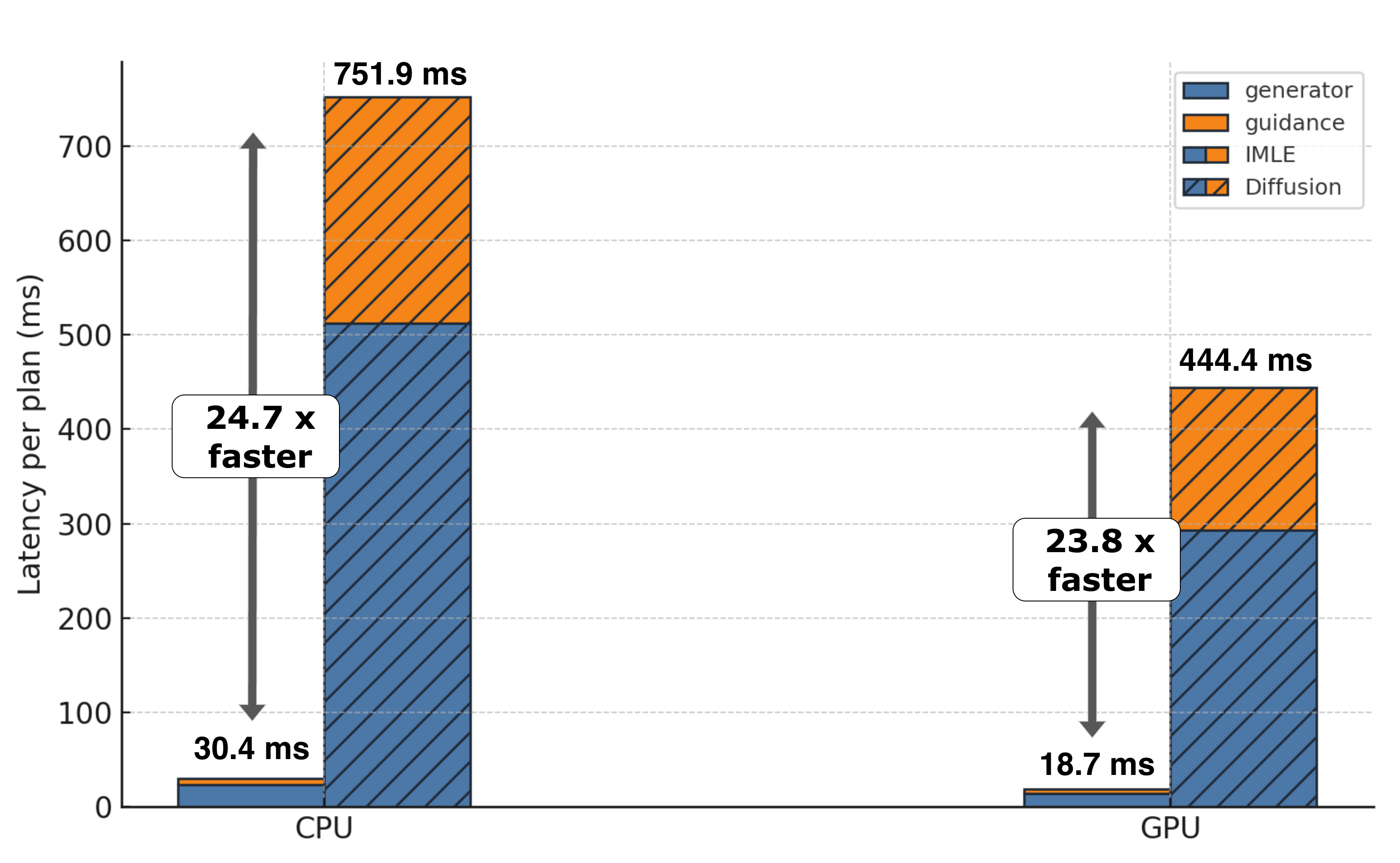}
     \caption{\textbf{IMLE vs Diffusion.} Median per-plan latency (ms) split into generator and guidance on CPU/GPU, averaged over Walker/Hopper/HalfCheetah (Batch Size 64)}
    \label{fig:imlelatency}
\end{figure}

\subsubsection{MuJoCo}
We compare our IMLE-based approach with Diffuser \cite{janner2022diffuser}, using the same model architecture and experimental setup to provide a controlled comparison. Our method achieves competitive performance across most environments while sidestepping the iterative denoising process required by diffusion models (Table I). We additionally break down the per-plan latency of each planner (Fig. \ref{fig:imlelatency}), separating generator and guidance times, since diffusion requires forward passes through both the generator and learned reward function that provides the guidance signal at each denoising step.

\begin{table}[ht]
\centering
\scriptsize
\setlength{\tabcolsep}{3pt}
\renewcommand{\arraystretch}{0.95}
\resizebox{\linewidth}{!}{%
\begin{tabular}{llcc}
\toprule
\textbf{Dataset} & \textbf{Environment} & \textbf{Diffuser}  & \textbf{IMLE+Exp RW} \\
\midrule
\textbf{Medium-Expert} & HalfCheetah  & 88.9 $\pm$ 0.3 & \textbf{91.9} $\pm$ 0.09\\
                       & Hopper & 103.3 $\pm$ 1.3 & \textbf{104.2}  $\pm$ 3.81\\
                       & Walker2d  & 106.9 $\pm$ 0.2 & \textbf{107.9} $\pm$ 0.40\\
\midrule
\textbf{Medium}        & HalfCheetah & 42.8 $\pm$ 0.3 & \textbf{43.1} $\pm$ 0.29\\
                       & Hopper & 74.3 $\pm$ 1.4 & \textbf{85.0} $\pm$ 4.02 \\
                       & Walker2d & \textbf{79.6} $\pm$ 0.55 & 78.3 $\pm$ 2.75\\
\midrule
\textbf{Medium-Replay} & HalfCheetah & 37.7 $\pm$ 0.5& \textbf{39.5} $\pm$ 0.59\\
                       & Hopper      & \textbf{93.6} $\pm$ 0.4 & 85.0 $\pm$ 4.02 \\
                       & Walker2d    & \textbf{70.6} $\pm$ 1.6 & 69.7 $\pm$ 3.84\\
\midrule
\multicolumn{2}{c}{\textbf{Average}} & 77.5 &  \textbf{78.47}\\
\midrule
\multicolumn{2}{c}\textbf{Sampling Frequency on CPU (Hz)} & 1.33 & \textbf{32.87} \\
\midrule
\multicolumn{2}{c}\textbf{Sampling Frequency on GPU (Hz)} & 2.25 & \textbf{53.52} \\
\bottomrule
\end{tabular}
}
\caption{Performance comparison across MuJoCo locomotion datasets. (batch size 64)}
\end{table}

\paragraph{Reward Weighting Impact} 
We analyze how reward weighting affects performance across datasets with varying trajectory quality (Table II). As expected, Medium-Expert shows modest gains since expert trajectories already bias toward high returns. The largest improvements occur in Medium-Replay and Medium datasets, where suboptimal behavior policies create low-return trajectory distributions. The superior performance of exponential weighting aligns with CAI theory, where optimal distributions have a Boltzmann form.

\begin{table}[ht]
\centering
\scriptsize
\setlength{\tabcolsep}{3pt}
\renewcommand{\arraystretch}{0.95}
\resizebox{\linewidth}{!}{%
\begin{tabular}{llccc}
\toprule
\textbf{Dataset} & \textbf{Environment} & \textbf{No RW} & \textbf{Lin RW} & \textbf{Exp RW}\\
\midrule
\textbf{Medium-Expert} & HalfCheetah  & 87.8 $\pm$ 0.94 & 90.6
    $\pm$ 0.26 & \textbf{91.9} $\pm$ 0.09\\
   & Hopper & 102.7 $\pm$ 4.07 & \textbf{111.2} $\pm$ 0.38 & 104.2  $\pm$ 3.81\\
   & Walker2d  & 108.8 $\pm$ 0.05 & \textbf{109.0} $\pm$ 0.04 & 107.9 $\pm$ 0.40\\
\midrule
\textbf{Medium}        & HalfCheetah & 35.5 $\pm$ 0.86 & \textbf{44.0} $\pm$ 0.09 & 43.1 $\pm$ 0.29\\
                       & Hopper  & 71.1 $\pm$ 5.39 & 65.0 $\pm$ 4.33 & \textbf{85.0} $\pm$ 4.02 \\
                       & Walker2d & 74.4 $\pm$ 3.23 & \textbf{81.6} $\pm$ 0.58 & 78.3 $\pm$ 2.75\\
\midrule
\textbf{Medium-Replay} & HalfCheetah  & 39.0 $\pm$ 0.62 & 38.0 $\pm$ 0.69 & \textbf{39.5} $\pm$  0.59\\
                       & Hopper       & 85.4 $\pm$ 5.51 & 75.9 $\pm$ 6.37  & \textbf{85.0} $\pm$ 4.02\\
                       & Walker2d    & 21.8 $\pm$ 6.57 & 30.5 $\pm$ 8.89 & \textbf{69.7} $\pm$ 3.84\\
\midrule
\multicolumn{2}{c}{\textbf{Average}} & 66.9 &  71.6 & \textbf{78.47}\\
\bottomrule
\end{tabular}
}
\caption{Performance comparison of reward-weighted IMLE variants.}
\end{table}

\subsubsection{Maze2D}
A key advantage of diffusion-based planners over their policy variants is their ability to perform well in sparse reward settings \cite{lu2025makes}. We observe that our IMLE-based planner achieves comparable performance on Maze2D, while providing a substantial speedup (Table III). Since all trajectories in this dataset are goal-reaching by construction, no reward weighting is necessary.

\begin{table}[h]
\centering
\resizebox{\linewidth}{!}{%
\begin{tabular}{llcc}
\toprule
\textbf{Dataset} & \textbf{Environment} & \textbf{Diffuser}  & \textbf{IMLE} \\
\midrule
\textbf{Single Task} & U-Maze  & 113.9 $\pm$ 3.1 & \textbf{124.8} $\pm$ 0.65 \\
                       & Medium & \textbf{121.5} $\pm$ 2.7 &  117.3 $\pm$ 3.53 \\
                       & Large  & 123.0 $\pm$ 6.4 & \textbf{129.2} $\pm$ 4.89 \\
\midrule
\multicolumn{2}{c}{\textbf{Average}} & 119.5 & \textbf{123.7} \\
\midrule
\textbf{Multi Task}        & U-Maze & 128.9 $\pm$ 1.8 & \textbf{132.3} $\pm$ 0.97 \\
                       & Medium & 127.2 $\pm$ 3.4 & \textbf{127.8} $\pm$ 2.60 \\
                       & Large & 132.1 $\pm$ 5.8 & \textbf{137.1} $\pm$ 4.41\\
\midrule
\multicolumn{2}{c}{\textbf{Average}} & 129.4 &  \textbf{132.4}\\
\midrule
\multicolumn{2}{c}{Sampling Frequency on CPU (Hz)} & 0.96 & \textbf{114.63} \\
\midrule
\multicolumn{2}{c}{Sampling Frequency on GPU (Hz)} & 1.37 & \textbf{101.28} \\
\bottomrule
\end{tabular}
}
\caption{Performance comparison across Maze2D Datasets. \\ (Batch Size 1)}
\end{table}

We additionally implemented our method in JAX, which achieves higher GPU sampling throughput than PyTorch (87.56 vs.\ 53.52 Hz on Locomotion and 133.98 vs.\ 101.28 Hz on Maze2D).

\subsection{Real-Time Navigation}
\subsubsection{Simulation}
We compare with CoBL \cite{mizuta2024cobl}, a DDIM sampler with 50 sampling steps, and an adapted Conditional Flow Matching (CFM) model \cite{lipman2022flow} using 9 ODE integration steps. Both generative baselines employ the same guidance function with 10 guidance iterations per sampling step.

\begin{table}[h]
\centering
\resizebox{\linewidth}{!}{
\begin{tabular}{lccccc}
\toprule
\textbf{Metric} & \textbf{MPPI} & \textbf{CoBL} & \textbf{CFM} & \textbf{IMLE} & \textbf{IMLE+MPPI} \\
\midrule
\multicolumn{6}{c}{\textbf{Collision Radius = 0.5 m}} \\
\midrule
Collision Rate (\%) $\downarrow$ & 10.00 & 10.20 & \underline{5.80} & 9.80 & \textbf{4.60} \\
Goal Error (m) $\downarrow$ & 0.521 & \textbf{0.050} & 0.431 & \underline{0.181} & 0.360 \\
Smoothness (m/s) $\downarrow$ & 0.772 & 0.606 & 0.623 & \underline{0.397} & \textbf{0.394} \\
Jerk (m/s$^3$) $\downarrow$ & 1.746 & 1.407 & 0.812 & \textbf{0.458} & \underline{0.479} \\
\midrule
\multicolumn{6}{c}{\textbf{Collision Radius = 0.7 m}} \\
\midrule
Collision Rate (\%) $\downarrow$ & \textbf{16.40} & 22.20 & 28.40 & \underline{20.60} & --- \\
Goal Error (m) $\downarrow$ & 0.570 & \textbf{0.048} & \underline{0.073} & 0.186 & --- \\
Smoothness (m/s) $\downarrow$ & 0.760 & \underline{0.605} & 0.612 & \textbf{0.394} & --- \\
Jerk (m/s$^3$) $\downarrow$ & 1.781 & 1.396 & \underline{0.740} & \textbf{0.461} & --- \\
\midrule
Sampling Frequency on CPU (Hz) $\uparrow$ & \textbf{125.00} & 0.41 & 2.65  & \underline{76.92} & 52.63 \\
\midrule
Sampling Frequency on GPU (Hz) $\uparrow$ & \textbf{142.86} & 0.71 & 4.30 & \underline{111.11} & 83.33\\
\bottomrule
\end{tabular}}
\caption{Planner performance across UCY Scenes. (batch size 64)}
\end{table}

With a collision radius of $0.5\,\mathrm{m}$, IMLE and IMLE+MPPI yield smoother, lower-jerk trajectories, consistent with their ability to capture realistic motion patterns from the dataset (Table IV). Because UCY pedestrians are well approximated by constant-velocity motion, objective-based planners (MPPI, CoBL, and CFM) more readily satisfy short-horizon safety constraints. Warm-starting MPPI with IMLE proposals improves constraint satisfaction while preserving trajectory smoothness.

At $0.7\,\mathrm{m}$, the mismatch between the safety constraint and the training distribution increases, as few trajectories maintain this level of clearance (Table IV). In this regime, generative models are most effective as proposal distributions for downstream optimization.

\begin{table}[h]
\centering
\scriptsize
\setlength{\tabcolsep}{3pt}
\renewcommand{\arraystretch}{0.95}
\resizebox{\linewidth}{!}{%
\begin{tabular}{lcccc}
\toprule
\textbf{Metric} & \textbf{Line} & \textbf{CoBL} & \textbf{CFM} & \textbf{IMLE} \\
\midrule
\multicolumn{5}{c}{\textbf{Collision Radius = 0.7 m}} \\
\midrule
Collision Rate (\%) $\downarrow$   & 16.40 & \textbf{8.80} & 11.40 & \underline{10.20} \\
Goal Error (m) $\downarrow$   & 0.570 & \underline{0.431} & 0.447 & \textbf{0.396} \\
Smoothness (m/s) $\downarrow$ & 0.760 & 0.760 & \underline{0.613} & \textbf{0.394} \\
Jerk (m/s$^3$) $\downarrow$   & 1.746 & \underline{0.759} & 0.787 & \textbf{0.484} \\
\midrule
Sampling Frequency on CPU (Hz) $\uparrow$ & \textbf{125.00} & 0.41 & 2.55 & \underline{52.63} \\
\midrule
Sampling Frequency on GPU (Hz) $\uparrow$ & \textbf{142.86} & 0.71 & 4.30 & \underline{83.33} \\
\bottomrule
\end{tabular}
}
\caption{Performance comparison between different warm-start strategies for MPPI across UCY scenes. (batch size 64)}
\end{table}

Warm-start strategies for MPPI show that while generative models alone often violate safety margins, using them as proposal distributions significantly improves performance (Table V). In particular, IMLE provides high-quality trajectories while maintaining real-time sampling.

\subsubsection{Mobile Robot}
We deploy the planner onboard a mobile robot with a high-frequency low-level controller, replanning at up to 50\,Hz on the onboard CPU using a batch of 8 trajectories per step. We evaluate real-world navigation with one to four pedestrians moving freely in a shared indoor environment with unknown goals, observing only past positions. The planner continuously updates its trajectory distribution to maintain collision avoidance while progressing toward the goal.

\begin{figure*}[t]
    \centering
    \includegraphics[width=\textwidth]{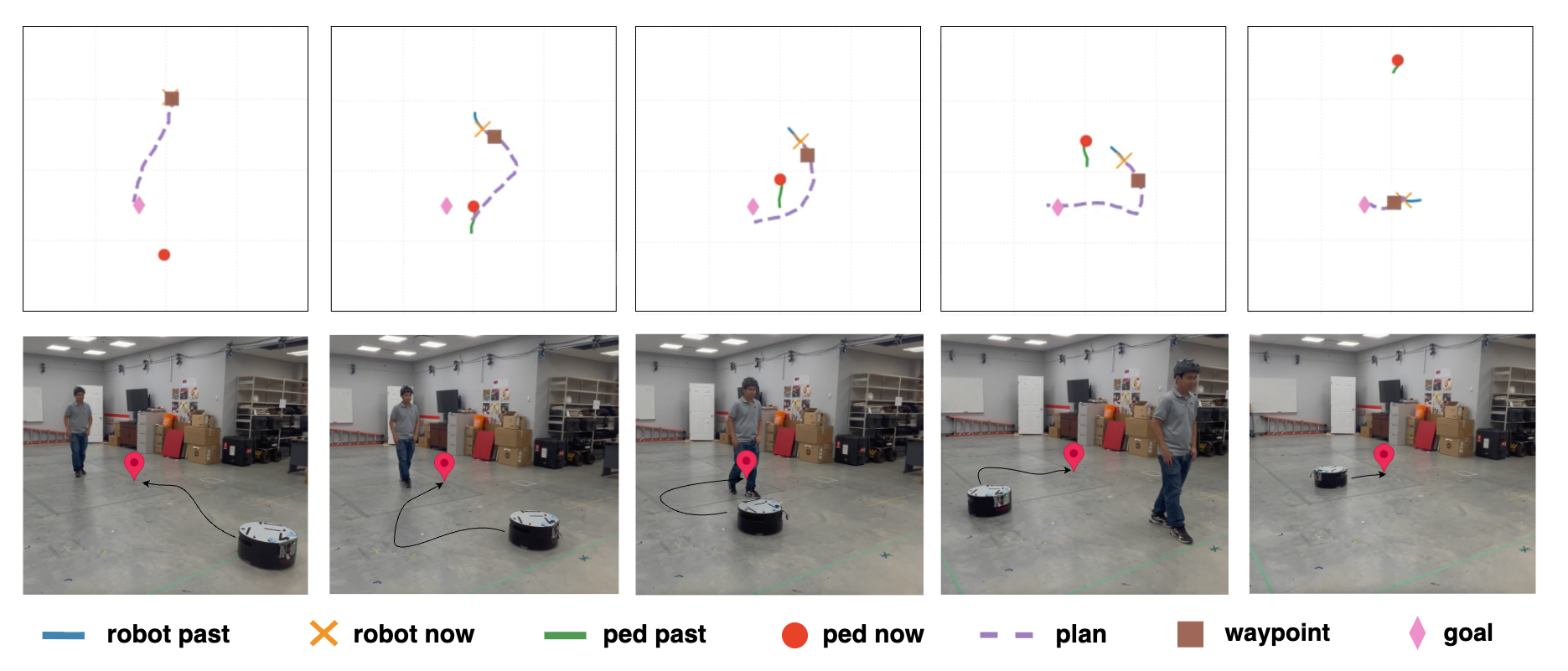}
    \caption{Using IMLE-generated plans at 50 Hz, the robot reaches the goal while avoiding collisions. The top row shows the conditioning variables (robot and pedestrian past and current states). We also plot the generated plan and the waypoint tracked by the low-level controller.} 
    \label{fig:robotnik_experiment}
\end{figure*}

\section{CONCLUSIONS AND FUTURE WORK}
We introduced a generative planning framework that adapts Implicit Maximum Likelihood Estimation (IMLE) for real-time MPC. By combining conditional IMLE with reward weighting, the method enables single-shot trajectory generation that preserves mode coverage while biasing toward high-return behaviors. Across offline RL benchmarks and real-time navigation tasks, IMLE achieves competitive performance with orders-of-magnitude faster inference than diffusion-based planners.

Several directions remain for future work. First, evaluating IMLE in data-scarce regimes is promising, as diffusion models may struggle under limited data due to the isotropic Gaussian forward process~\cite{vashist2024rejection}, while IMLE has demonstrated stronger data efficiency in policy learning~\cite{rana2025imle}. Second, the coverage properties of IMLE may support uncertainty-aware planning in dynamic environments, as diverse trajectory proposals can improve downstream optimization and safety filtering~\cite{wagenmaker2025posterior}. Finally, like other generative models, IMLE degrades when optimal trajectories lie outside training data; while augmentation helps, adaptive mechanisms~\cite{liang2023adaptdiffuser} may further improve robustness.



\section*{ACKNOWLEDGMENT}

This research was enabled in part by support from NSERC, the BC DRI Group, and the Digital Research Alliance of Canada. The authors thank Chirag Vashist and Shichong Peng for insightful discussions on IMLE, and Kurtis Yang for assistance with real-world experimental setup.


\printbibliography

@article{kappen2005path,
    title={Path integrals and symmetry breaking for optimal control theory},
    author={Kappen, Hilbert J},
    journal={Journal of statistical mechanics: theory and experiment},
    volume={2005},
    number={11},
    pages={P11011},
    year={2005},
    publisher={IOP Publishing}
}

@article{goodfellow2014generative,
    title={Generative adversarial nets},
    author={Goodfellow, Ian J and Pouget-Abadie, Jean and Mirza, Mehdi and Xu, Bing and Warde-Farley, David and Ozair, Sherjil and Courville, Aaron and Bengio, Yoshua},
    journal={Advances in Neural Information Processing Systems},
    volume={27},
    year={2014}
}

@inproceedings{emmons2022rvs,
    title={RvS: What is Essential for Offline {RL} via Supervised Learning?},
    author={Emmons, Scott and Eysenbach, Benjamin and Kostrikov, Ilya and Levine, Sergey},
    booktitle={International Conference on Learning Representations},
    year={2022},
}

@article{schmidhuber2019reinforcement,
  title={Reinforcement Learning Upside Down: Don't Predict Rewards--Just Map Them to Actions},
  author={Schmidhuber, Juergen},
  journal={arXiv preprint arXiv:1912.02875},
  year={2019}
}

@article{chen2021decision,
  title={Decision transformer: Reinforcement learning via sequence modeling},
  author={Chen, Lili and Lu, Kevin and Rajeswaran, Aravind and Lee, Kimin and Grover, Aditya and Laskin, Misha and Abbeel, Pieter and Srinivas, Aravind and Mordatch, Igor},
  journal={Advances in Neural Information Processing Systems},
  volume={34},
  pages={15084--15097},
  year={2021}
}

@inproceedings{janner2022diffuser,
    title = {Planning with Diffusion for Flexible Behavior Synthesis},
    author = {Janner, Michael and Du, Yilun and Tenenbaum, Joshua and Levine, Sergey},
    booktitle = {International Conference on Machine Learning},
    year = {2022},
}

@inproceedings{janner2021sequence,
  title = {Offline Reinforcement Learning as One Big Sequence Modeling Problem},
  author = {Janner, Michael and Li, Qiyang and Levine, Sergey},
  booktitle = {Advances in Neural Information Processing Systems},
  year = {2021},
}

@article{dong2024diffuserlite,
  title={Diffuserlite: Towards real-time diffusion planning},
  author={Dong, Zibin and Hao, Jianye and Yuan, Yifu and Ni, Fei and Wang, Yitian and Li, Pengyi and Zheng, Yan},
  journal={Advances in Neural Information Processing Systems},
  volume={37},
  pages={122556--122583},
  year={2024}
}

@inproceedings{lu2025habitizing,
    title={Habitizing Diffusion Planning for Efficient and Effective Decision Making},
    author={Lu, Haofei and Shen, Yifei and Li, Dongsheng and Xing, Junliang and Han, Dongqi},
    booktitle={Forty-second International Conference on Machine Learning},
    year={2025},
}

@inproceedings{mizuta2024cobl,
  title={Cobl-diffusion: Diffusion-based conditional robot planning in dynamic environments using control barrier and lyapunov functions},
  author={Mizuta, Kazuki and Leung, Karen},
  booktitle={2024 IEEE/RSJ International Conference on Intelligent Robots and Systems (IROS)},
  pages={13801--13808},
  year={2024},
  organization={IEEE}
}

@article{mizuta2025unified,
  title={Unified Generation-Refinement Planning: Bridging Flow Matching and Sampling-Based MPC},
  author={Mizuta, Kazuki and Leung, Karen},
  journal={arXiv preprint arXiv:2508.01192},
  year={2025}
}

@article{todorov2009compositionality,
  title={Compositionality of optimal control laws},
  author={Todorov, Emanuel},
  journal={Advances in Neural Information Processing Systems},
  volume={22},
  year={2009}
}

@inproceedings{ajay2023is,
title={Is Conditional Generative Modeling all you need for Decision Making?},
author={Ajay, Anurag and Du, Yilun and Gupta, Abhi and Tenenbaum, Joshua B. and Jaakkola, Tommi S. and Agrawal, Pulkit},
booktitle={The Eleventh International Conference on Learning Representations },
year={2023},
}

@inproceedings{perez2018film,
  title={Film: Visual reasoning with a general conditioning layer},
  author={Perez, Ethan and Strub, Florian and De Vries, Harm and Dumoulin, Vincent and Courville, Aaron},
  booktitle={Proceedings of the AAAI conference on artificial intelligence},
  volume={32},
  number={1},
  year={2018}
}

@article{chi2023diffusion,
  title={Diffusion policy: Visuomotor policy learning via action diffusion},
  author={Chi, Cheng and Xu, Zhenjia and Feng, Siyuan and Cousineau, Eric and Du, Yilun and Burchfiel, Benjamin and Tedrake, Russ and Song, Shuran},
  journal={The International Journal of Robotics Research},
  pages={02783649241273668},
  year={2023},
  publisher={SAGE Publications Sage UK: London, England}
}

@article{peng2019advantage,
  title={Advantage-weighted regression: Simple and scalable off-policy reinforcement learning},
  author={Peng, Xue Bin and Kumar, Aviral and Zhang, Grace and Levine, Sergey},
  journal={arXiv preprint arXiv:1910.00177},
  year={2019}
}

@Inproceedings{ivanovic2023trajdata,
  author = {Ivanovic, Boris and Song, Guanyu and Gilitschenski, Igor and Pavone, Marco},
  title = {{trajdata}: A Unified Interface to Multiple Human Trajectory Datasets},
  booktitle = {Proceedings of the Neural Information Processing Systems (NeurIPS) Track on Datasets and Benchmarks},
  month = dec,
  year = {2023},
  address = {New Orleans, USA},
}

@INPROCEEDINGS{Pellegrini2009youll,
  author={Pellegrini, Stefano and Ess, Andreas and Schindler, Konrad and Van Gool, Luc},
  booktitle={2009 IEEE 12th International Conference on Computer Vision}, 
  title={You'll never walk alone: Modeling social behavior for multi-target tracking}, 
  year={2009},
  pages={261-268}
}

@article{Lerner2007crowds,
    author = {Lerner, Alon and Chrysanthou, Yiorgos and Lischinski, Dani},
    title = {Crowds by Example},
    journal = {Computer Graphics Forum},
    volume = {26},
    number = {3},
    pages = {655-664},
    year = {2007}
}

@article{ho2020denoising,
  title={Denoising diffusion probabilistic models},
  author={Ho, Jonathan and Jain, Ajay and Abbeel, Pieter},
  journal={Advances in Neural Information Processing Systems},
  volume={33},
  pages={6840--6851},
  year={2020}
}

@inproceedings{xiao2023safediffuser,
  title={Safediffuser: Safe planning with diffusion probabilistic models},
  author={Xiao, Wei and Wang, Tsun-Hsuan and Gan, Chuang and Hasani, Ramin and Lechner, Mathias and Rus, Daniela},
  booktitle={The Thirteenth International Conference on Learning Representations},
  year={2023}
}

@article{levine2018reinforcement,
  title={Reinforcement learning and control as probabilistic inference: Tutorial and review},
  author={Levine, Sergey},
  journal={arXiv preprint arXiv:1805.00909},
  year={2018}
}

@book{ziebart2010modeling,
  title={Modeling purposeful adaptive behavior with the principle of maximum causal entropy},
  author={Ziebart, Brian D},
  year={2010},
  publisher={Carnegie Mellon University}
}

@inproceedings{
lu2025makes,
title={What Makes a Good Diffusion Planner for Decision Making?},
author={Haofei Lu and Dongqi Han and Yifei Shen and Dongsheng Li},
booktitle={The Thirteenth International Conference on Learning Representations},
year={2025},
}

@inproceedings{williams2016aggressive,
  title={Aggressive driving with model predictive path integral control},
  author={Williams, Grady and Drews, Paul and Goldfain, Brian and Rehg, James M and Theodorou, Evangelos A},
  booktitle={2016 IEEE international Conference on Robotics and Automation (ICRA)},
  pages={1433--1440},
  year={2016},
  organization={IEEE}
}

@inproceedings{lu2023contrastive,
  title={Contrastive energy prediction for exact energy-guided diffusion sampling in offline reinforcement learning},
  author={Lu, Cheng and Chen, Huayu and Chen, Jianfei and Su, Hang and Li, Chongxuan and Zhu, Jun},
  booktitle={International Conference on Machine Learning},
  pages={22825--22855},
  year={2023},
  organization={PMLR}
}

@inproceedings{feng2025on,
title={On the Guidance of Flow Matching},
author={Feng, Ruiqi and Yu, Chenglei and Deng, Wenhao and Hu, Peiyan and Wu, Tailin},
booktitle={Forty-second International Conference on Machine Learning},
year={2025},
}

@inproceedings{rana2025imle, 
    title={IMLE Policy: Fast and Sample Efficient Visuomotor Policy Learning via Implicit Maximum Likelihood Estimation}, 
    author={Rana, Krishan and Lee, Robert and Pershouse, David and Suenderhauf, Niko}, 
    booktitle={Proceedings of Robotics: Science and Systems (RSS)}, 
    year= {2025}
    }

@inproceedings{liang2023adaptdiffuser,
  title={AdaptDiffuser: Diffusion Models as Adaptive Self-evolving Planners},
  author={Liang, Zhixuan and Mu, Yao and Ding, Mingyu and Ni, Fei and Tomizuka, Masayoshi and Luo, Ping},
  booktitle={International Conference on Machine Learning},
  pages={20725--20745},
  year={2023},
  organization={PMLR}
}

@inproceedings{nguyen2016exponential,
  title={Exponential control barrier functions for enforcing high relative-degree safety-critical constraints},
  author={Nguyen, Quan and Sreenath, Koushil},
  booktitle={2016 American Control Conference (ACC)},
  pages={322--328},
  year={2016},
  organization={IEEE}
}

@INPROCEEDINGS{CBForiginal,
  author={Ames, Aaron D. and Coogan, Samuel and Egerstedt, Magnus and Notomista, Gennaro and Sreenath, Koushil and Tabuada, Paulo},
  booktitle={2019 18th European Control Conference (ECC)}, 
  title={Control Barrier Functions: Theory and Applications}, 
  year={2019},
  volume={},
  number={},
  pages={3420-3431},
  doi={10.23919/ECC.2019.8796030}}

@misc{li2018implicit,
      title={Implicit Maximum Likelihood Estimation}, 
      author={Li, Ke and Malik, Jitendra},
      year={2018},
      eprint={1809.09087},
      archivePrefix={arXiv},
      primaryClass={cs.LG},
      url={https://arxiv.org/abs/1809.09087}, 
}

@inproceedings{vashist2024rejection,
  title={Rejection sampling imle: Designing priors for better few-shot image synthesis},
  author={Vashist, Chirag and Peng, Shichong and Li, Ke},
  booktitle={European Conference on Computer Vision},
  pages={441--456},
  year={2024},
  organization={Springer}
}

@inproceedings{Aghabozorgi2023AdaptiveIF,
  title={Adaptive IMLE for Few-shot Pretraining-free Generative Modelling},
  author={Mehran Aghabozorgi and Shichong Peng and Ke Li},
  booktitle={International Conference on Machine Learning},
  year={2023},
  url={https://api.semanticscholar.org/CorpusID:260810286}
}

@article{dhariwal2021diffusion,
  title={Diffusion models beat gans on image synthesis},
  author={Dhariwal, Prafulla and Nichol, Alexander},
  journal={Advances in neural information processing systems},
  volume={34},
  pages={8780--8794},
  year={2021}
}

@article{li2020multimodal,
  title={Multimodal image synthesis with conditional implicit maximum likelihood estimation},
  author={Li, Ke and Peng, Shichong and Zhang, Tianhao and Malik, Jitendra},
  journal={International Journal of Computer Vision},
  volume={128},
  number={10},
  pages={2607--2628},
  year={2020},
  publisher={Springer}
}

@article{peng2022chimle,
  title={Chimle: Conditional hierarchical imle for multimodal conditional image synthesis},
  author={Peng, Shichong and Moazenipourasil, Seyed Alireza and Li, Ke},
  journal={Advances in Neural Information Processing Systems},
  volume={35},
  pages={280--296},
  year={2022}
}

@article{fu2020d4rl,
  title={D4rl: Datasets for deep data-driven reinforcement learning},
  author={Fu, Justin and Kumar, Aviral and Nachum, Ofir and Tucker, George and Levine, Sergey},
  journal={arXiv preprint arXiv:2004.07219},
  year={2020}
}

@article{wagenmaker2025posterior,
  title={Posterior Behavioral Cloning: Pretraining BC Policies for Efficient RL Finetuning},
  author={Wagenmaker, Andrew and Dong, Perry and Tsao, Raymond and Finn, Chelsea and Levine, Sergey},
  journal={arXiv preprint arXiv:2512.16911},
  year={2025}
}

@article{kumar2020conservative,
  title={Conservative q-learning for offline reinforcement learning},
  author={Kumar, Aviral and Zhou, Aurick and Tucker, George and Levine, Sergey},
  journal={Advances in neural information processing systems},
  volume={33},
  pages={1179--1191},
  year={2020}
}

@article{janner2021offline,
  title={Offline reinforcement learning as one big sequence modeling problem},
  author={Janner, Michael and Li, Qiyang and Levine, Sergey},
  journal={Advances in neural information processing systems},
  volume={34},
  pages={1273--1286},
  year={2021}
}

@article{kostrikov2021offline,
  title={Offline reinforcement learning with implicit q-learning},
  author={Kostrikov, Ilya and Nair, Ashvin and Levine, Sergey},
  journal={arXiv preprint arXiv:2110.06169},
  year={2021}
}

@article{yu2020mopo,
  title={Mopo: Model-based offline policy optimization},
  author={Yu, Tianhe and Thomas, Garrett and Yu, Lantao and Ermon, Stefano and Zou, James Y and Levine, Sergey and Finn, Chelsea and Ma, Tengyu},
  journal={Advances in neural information processing systems},
  volume={33},
  pages={14129--14142},
  year={2020}
}

@article{kidambi2020morel,
  title={Morel: Model-based offline reinforcement learning},
  author={Kidambi, Rahul and Rajeswaran, Aravind and Netrapalli, Praneeth and Joachims, Thorsten},
  journal={Advances in neural information processing systems},
  volume={33},
  pages={21810--21823},
  year={2020}
}

@article{argenson2020model,
  title={Model-based offline planning},
  author={Argenson, Arthur and Dulac-Arnold, Gabriel},
  journal={arXiv preprint arXiv:2008.05556},
  year={2020}
}

@article{lipman2022flow,
  title={Flow matching for generative modeling},
  author={Lipman, Yaron and Chen, Ricky TQ and Ben-Hamu, Heli and Nickel, Maximilian and Le, Matt},
  journal={arXiv preprint arXiv:2210.02747},
  year={2022}
}

\clearpage

\FloatBarrier

\section*{APPENDIX}

For reproducibility, we summarize the key architectural, training, and planning hyperparameters used in our experiments. These include the primary settings that influence model capacity, training dynamics, and planning performance.

\setlength{\tabcolsep}{4pt}
\renewcommand{\arraystretch}{1.05}


\section{Implementation Details}
\label{sec:appendix}

\FloatBarrier

\subsection{D4RL Locomotion}
\label{subsec:impl_locomotion}

\begin{table}[h]
\centering
\caption{Key hyperparameters for D4RL locomotion experiments.}
\label{tab:locomotion}
\small
\begin{tabular}{lc}
\toprule
Parameter & Value \\
\midrule
Latent dimension & 16 \\
Channel multipliers & (1,4,16) \\
Planning horizon & 32 \\
IMLE sample factor $K$ & 4 \\
\midrule
Learning rate & $1\times10^{-3}$ \\
Batch size & 32 \\
Training steps & 100k \\
Loss & L2 \\
\bottomrule
\end{tabular}
\end{table}

\subsection{D4RL Maze2D}
\label{subsec:impl_maze2d}

\begin{table}[h]
\centering
\caption{Key hyperparameters for D4RL Maze2D experiments.}
\label{tab:maze2d}
\small
\begin{tabular}{lc}
\toprule
Parameter & Value \\
\midrule
Latent dimension & 3 \\
Channel multipliers & (1,2,4,8) \\
Planning horizon & 256 \\
IMLE sample factor $K$ & 10 \\
\midrule
Learning rate & $1\times10^{-4}$ \\
Batch size & 32 \\
Training steps & 100k \\
Loss & L2 \\
\bottomrule
\end{tabular}
\end{table}

\subsection{Pedestrian Navigation (ETH/UCY)}
\label{subsec:impl_navigation}

We report three horizon presets: corresponding to planning horizons of 20, 40, and 80 steps with time discretizations of $0.4$, $0.2$, and $0.1$ seconds respectively.

\begin{table}[H]
\centering
\caption{Architecture and training hyperparameters for navigation experiments.}
\label{tab:nav_hyperparams}
\small
\setlength{\tabcolsep}{4pt}
\renewcommand{\arraystretch}{1.05}
\begin{tabular}{lccc}
\toprule
 & IMLE & Diffusion & CFM \\
\midrule
\multicolumn{4}{c}{\textbf{Architecture}} \\
\midrule
Base dimension & 32 & 32 & 32 \\
Channel multipliers & (1,4)/(1,2,4)/(1,2,4,8) & same & same \\
\midrule
\multicolumn{4}{c}{\textbf{Training}} \\
\midrule
Learning rate & $1\times10^{-4}$ & $2\times10^{-5}$ & $2\times10^{-5}$ \\
Batch size & 64 & 64 & 64 \\
Training steps & 250k & 1M & 1M \\
Loss & L2 & L1 & L1 \\
\bottomrule
\end{tabular}
\end{table}

\begin{table}[h]
\centering
\caption{Planning and sampling parameters.}
\label{tab:planning}
\small
\begin{tabular}{lc}
\toprule
Parameter & Value \\
\midrule
Planning horizon & 20 / 40 / 80 \\
Time discretization $\Delta t$ & 0.4 / 0.2 / 0.1 s \\
Trajectory samples & 64 \\
MPPI perturbation samples & 32 \\
MPPI temperature & 0.5 \\
Safety radius & 0.5 / 0.7 m \\
DDIM steps & 50 \\
CFM integration steps & 10 \\
$\tau$ schedule &
\begin{tabular}[t]{@{}l@{}}
[0.0, 0.5, 0.8, 0.85, 0.9,\\
0.92, 0.94, 0.96, 0.98, 1.0]
\end{tabular} \\
\bottomrule
\end{tabular}
\end{table}



\section{Additional Quantitative Results}
\label{sec:additional_results}

This section provides additional experimental comparisons across Locomotion, Maze2D, and UCY benchmarks.

\subsection{UCY Additional Ablations}

Tables~\ref{tab:nav40} and~\ref{tab:nav80} report warm-start performance across 500 UCY scenes for horizons $H=40$ and $H=80$. IMLE remains competitive for warm-starting MPPI, achieving lower goal error and competitive collision rates while producing smoother trajectories than a straight-line initialization.

At finer discretizations and longer horizons, diffusion-based methods (CoBL and CFM) become more competitive. The higher-dimensional trajectory space allows guidance to adjust trajectories more gradually to satisfy reward and safety constraints. This improves sample quality but requires multiple sampling steps, whereas IMLE generates trajectories in a single forward pass.

\begin{table}[h]
\centering
\small
\setlength{\tabcolsep}{3pt}
\renewcommand{\arraystretch}{1.05}
\begin{tabular}{lcccc}
\toprule
Metric & Straight & CoBL & CFM & IMLE \\
\midrule
Collision Rate (\%) & 7.40 & 6.40 & 8.40 & \textbf{5.60} \\
Goal Error (m) & 0.829 & 0.913 & 0.842 & \textbf{0.561} \\
Smoothness (m/s) & 1.715 & 0.891 & \textbf{0.493} & 0.817 \\
Jerk (m/s$^3$) & 7.013 & \textbf{0.883} & 1.141 & 1.549 \\
CPU Sampling (Hz) & \textbf{90.91} & 0.19 & 1.50 & 34.48 \\
GPU Sampling (Hz) & \textbf{125} & 0.56 & 2.05 & 58.82 \\
\bottomrule
\end{tabular}
\caption{Performance comparison between warm starts across 500 UCY scenes (batch size 64, $H=40$).}
\label{tab:nav40}
\end{table}

\begin{table}[H]
\centering
\small
\setlength{\tabcolsep}{3pt}
\renewcommand{\arraystretch}{1.05}
\begin{tabular}{lcccc}
\toprule
Metric & Straight & CoBL & CFM & IMLE \\
\midrule
Collision Rate (\%) & 1.40 & \textbf{0.60} & 1.20 & 1.40 \\
Goal Error (m) & 0.518 & 0.672 & 0.580 & \textbf{0.352} \\
Smoothness (m/s) & 3.736 & 1.571 & \textbf{0.465} & 1.016 \\
Jerk (m/s$^3$) & 27.997 & 3.152 & \textbf{2.405} & 3.505 \\
CPU Sampling (Hz) & \textbf{50.00} & 0.10 & 0.70 & 21.74 \\
GPU Sampling (Hz) & \textbf{83.33} & 0.49 & 1.94 & 40.00 \\
\bottomrule
\end{tabular}
\caption{Performance comparison between warm starts across 500 UCY scenes (batch size 64, $H=80$).}
\label{tab:nav80}
\end{table}

\begin{table*}[t]
\centering
\small
\setlength{\tabcolsep}{3pt}
\renewcommand{\arraystretch}{1.05}
\begin{tabular}{llcccccccccc}
\toprule
\textbf{Dataset} & \textbf{Environment} 
& \textbf{BC} & \textbf{CQL} & \textbf{IQL} 
& \textbf{DT} & \textbf{TT}
& \textbf{MOPO} & \textbf{MOReL} & \textbf{MBOP}
& \textbf{Diffuser} & \textbf{IMLE+Exp RW} \\
\midrule
\textbf{Medium-Expert}
& HalfCheetah 
& 55.2 & 91.6 & 86.7 & 86.8 & 95.0 & 63.3 & 53.3 & \textbf{105.9}
& 88.9 $\pm$ 0.3 & 91.9 $\pm$ 0.09 \\
& Hopper 
& 52.5 & \textbf{105.4} & 91.5 & \textbf{107.6} & \textbf{110.0} & 23.7 & \textbf{108.7} & 55.1
& 103.3 $\pm$ 1.3 & 104.2 $\pm$ 3.81 \\
& Walker2d  
& \textbf{107.5} & \textbf{108.8} & \textbf{109.6} & \textbf{108.1} & 101.9 & 44.6 & 95.6 & 70.2
& \textbf{106.9} $\pm$ 0.2 & \textbf{107.9} $\pm$ 0.40 \\
\midrule
\textbf{Medium}
& HalfCheetah 
& 42.6 & 44.0 & \textbf{47.4} & 42.6 & \textbf{46.9} & 42.3 & 42.1 & 44.6
& 42.8 $\pm$ 0.3 & 43.1 $\pm$ 0.29 \\
& Hopper 
& 52.9 & 58.5 & 66.3 & 67.6 & 61.1 & 28.0 & \textbf{95.4} & 48.8
& 74.3 $\pm$ 1.4 & 85.0 $\pm$ 4.02 \\
& Walker2d 
& 75.3 & 72.5 & \textbf{78.3} & 74.0 & \textbf{79.0} & 17.8 & \textbf{77.8} & 41.0
& \textbf{79.6} $\pm$ 0.55 & \textbf{78.3} $\pm$ 2.75 \\
\midrule
\textbf{Medium-Replay}
& HalfCheetah 
& 36.6 & 45.5 & 44.2 & 36.6 & 41.9 & \textbf{53.1} & 40.2 & 42.3
& 37.7 $\pm$ 0.5 & 39.5 $\pm$ 0.59 \\
& Hopper      
& 18.1 & \textbf{95.0} & \textbf{94.7} & 82.7 & \textbf{91.5} & 67.5 & \textbf{93.6} & 12.4
& \textbf{93.6} $\pm$ 0.4 & 85.0 $\pm$ 4.02 \\
& Walker2d    
& 26.0 & 77.2 & 73.9 & 66.6 & \textbf{82.6} & 39.0 & 49.8 & 9.7
& 70.6 $\pm$ 1.6 & 69.7 $\pm$ 3.84 \\
\midrule
\multicolumn{2}{c}{\textbf{Average}} 
& 51.9 & \textbf{77.6} & \textbf{77.0} & 74.7 & \textbf{78.9} & 42.1 & 72.9 & 47.8
& \textbf{77.5} & \textbf{78.4} \\
\bottomrule
\end{tabular}
\caption{\centering
Performance comparison across MuJoCo locomotion datasets.
Baselines are taken from~\cite{janner2022diffuser}. 
Bold indicates within 5\% of the maximum reward.
}
\label{tab:locomotion_results}
\end{table*}

\subsection{Locomotion Additional Comparisons}

Table~\ref{tab:locomotion_results} compares IMLE with a range of offline reinforcement learning and model-based methods on the D4RL locomotion benchmarks \cite{kumar2020conservative, kostrikov2021offline, chen2021decision, janner2021offline, yu2020mopo, kidambi2020morel, argenson2020model}.

As expected, model-free offline RL algorithms such as CQL \cite{kumar2020conservative}, and IQL \cite{kostrikov2021offline} remain highly competitive on these tasks since locomotion environments primarily require policy learning rather than long-horizon planning. Nevertheless, IMLE achieves results within the top-performing methods across most environments.

\subsection{Maze2D Additional Comparisons}

Table~\ref{tab:maze_results} evaluates performance on Maze2D navigation tasks. Unlike locomotion, these environments contain sparse rewards and require long-horizon planning. As a result, trajectory generation methods significantly outperform policy-based offline RL approaches such as CQL and IQL.
\begin{table}[h]
\centering
\small
\setlength{\tabcolsep}{3pt}
\renewcommand{\arraystretch}{1.05}
\resizebox{\linewidth}{!}{%
\begin{tabular}{llccccc}
\toprule
\textbf{Setting} & \textbf{Env}
& \textbf{MPPI} & \textbf{CQL} & \textbf{IQL}
& \textbf{Diffuser} & \textbf{IMLE} \\
\midrule
\textbf{Single Task} & U-Maze
& 33.2 & 5.7 & 47.4
& 113.9 $\pm$ 3.1
& \textbf{124.8} $\pm$ 0.65 \\
& Medium
& 10.2 & 5.0 & 34.9
& \textbf{121.5} $\pm$ 2.7
& 117.3 $\pm$ 3.53 \\
& Large
& 5.1 & 12.5 & 58.6
& 123.0 $\pm$ 6.4
& \textbf{129.2} $\pm$ 4.89 \\
\midrule
\multicolumn{2}{c}{\textbf{Single-task Average}}
& 16.2 & 7.7 & 47.0
& 119.5
& \textbf{123.7} \\
\midrule
\textbf{Multi Task} & U-Maze
& 41.2 & --- & 24.8
& 128.9 $\pm$ 1.8
& \textbf{132.3} $\pm$ 0.97 \\
& Medium
& 15.4 & --- & 12.1
& 127.2 $\pm$ 3.4
& \textbf{127.8} $\pm$ 2.60 \\
& Large
& 8.0 & --- & 13.9
& 132.1 $\pm$ 5.8
& \textbf{137.1} $\pm$ 4.41 \\
\midrule
\multicolumn{2}{c}{\textbf{Multi-task Average}}
& 21.5 & --- & 16.9
& 129.4
& \textbf{132.4} \\
\bottomrule
\end{tabular}%
}
\caption{Performance comparison across Maze2D datasets (batch size 1). Baselines are taken from \cite{janner2022diffuser}.}
\label{tab:maze_results}
\end{table}

\section{JAX Implementation}

Table~\ref{tab:jax_speed} reports sampling throughput for our Torch and JAX implementations. 
JAX provides additional acceleration on GPU, particularly for larger batch sizes where 
XLA compilation improves parallel execution.

\begin{table}[h]
\centering
\scriptsize
\setlength{\tabcolsep}{3pt}
\renewcommand{\arraystretch}{0.95}
\begin{tabular}{lcccc}
\toprule
& \multicolumn{2}{c}{\textbf{Torch}} & \multicolumn{2}{c}{\textbf{JAX}} \\
\cmidrule(lr){2-3} \cmidrule(lr){4-5}
\textbf{Task} & CPU & GPU & CPU & GPU \\
\midrule
Locomotion & \textbf{32.87} & 53.52 & 19.23 & \textbf{87.56} \\
Maze       & \textbf{114.63} & 101.28 & 91.04 & \textbf{133.98} \\
\bottomrule
\end{tabular}
\caption{Sampling frequency (Hz) of Torch and JAX implementations.}
\label{tab:jax_speed}
\end{table}
\section{Additional Qualitative Results}

\begin{figure}[H]
\centering

\begin{minipage}{0.39\linewidth}
\centering
\includegraphics[width=\linewidth]{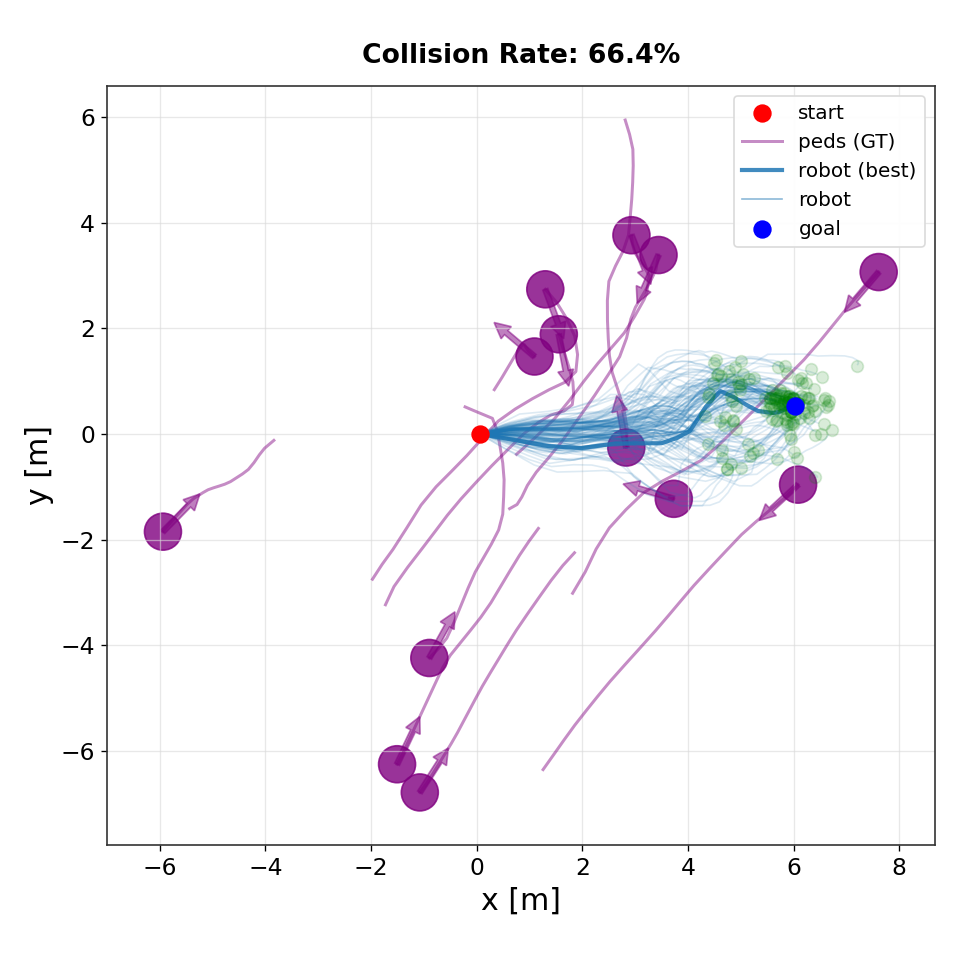}
\small (a) CoBL (80.5\% coll.)
\end{minipage}

\vspace{4pt}

\begin{minipage}{0.39\linewidth}
\centering
\includegraphics[width=\linewidth]{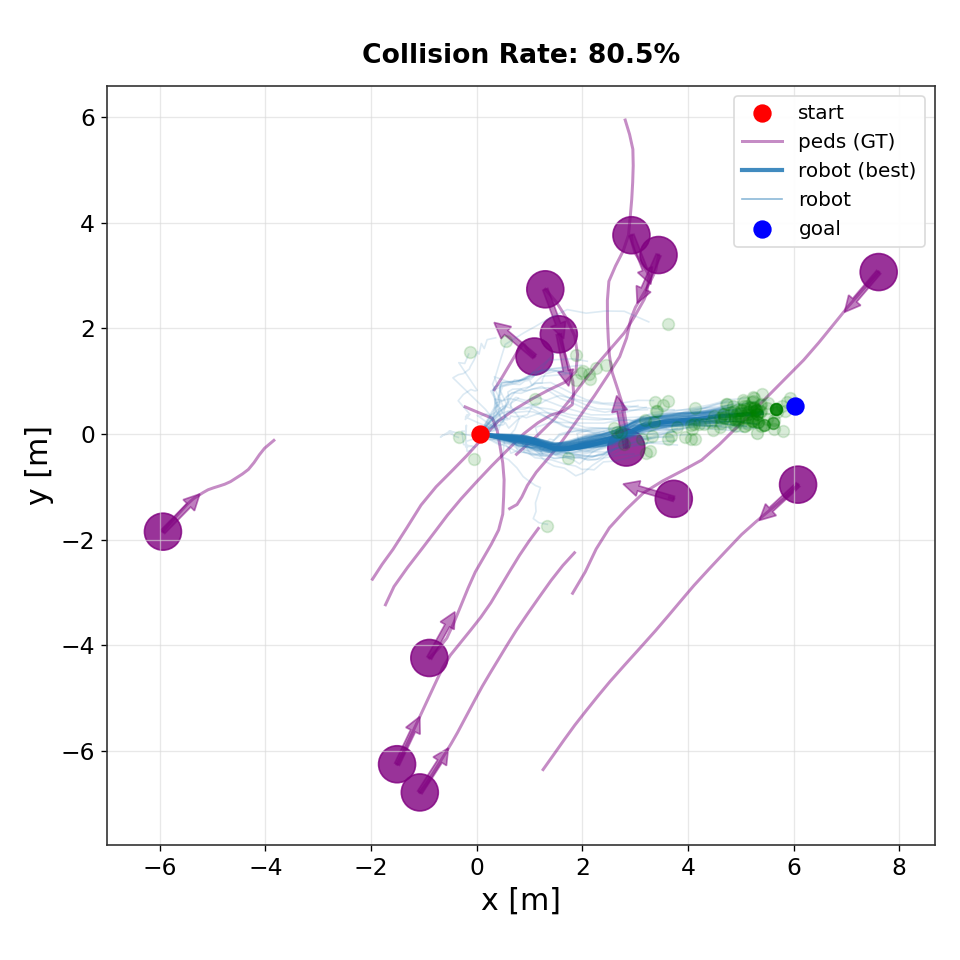}
\small (b) CFM (66.4\% coll.)
\end{minipage}

\vspace{4pt}

\begin{minipage}{0.39\linewidth}
\centering
\includegraphics[width=\linewidth]{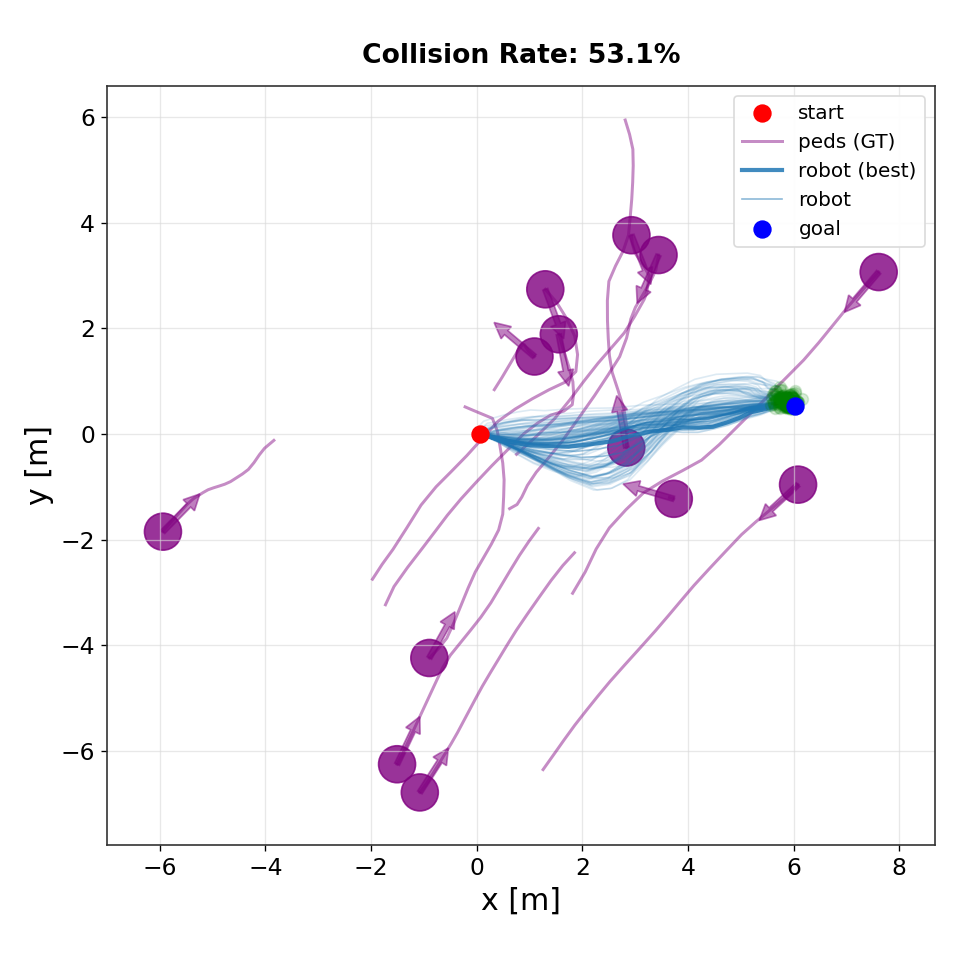}
\small (c) IMLE (53.1\% coll.)
\end{minipage}

\caption{
Qualitative comparison of trajectory proposals for a representative scene.
Each panel shows sampled trajectories and the selected plan.
}

\label{fig:appendix_qualitative}
\end{figure}
\end{document}